\documentclass[10pt,twocolumn,letterpaper]{article}

\usepackage{cvpr}
\usepackage{times}
\usepackage{epsfig}
\usepackage{graphicx}
\usepackage{amsmath}
\usepackage{amssymb}
\usepackage[inline]{enumitem}
\usepackage{siunitx}

\usepackage[pagebackref=true,breaklinks=true,letterpaper=true,colorlinks,bookmarks=false]{hyperref}

\cvprfinalcopy 

\def\cvprPaperID{7340} 

\ifcvprfinal\fi
\begin{document}

\title{KISS: Keeping It Simple for Scene Text Recognition}

\author{Christian Bartz, Joseph Bethge, Haojin Yang, and Christoph Meinel\\
Hasso-Plattner-Institute, University of Potsdam\\
Prof.-Dr.-Helmert Strasse 2-3, 14482 Potsdam, Germany\\
{\tt\small [christian.bartz, haojin.yang, joseph.bethge, christoph.meinel]@hpi.de}
}

\maketitle

\begin{abstract}
   \noindent
   Over the past few years, several new methods for scene text recognition have been proposed.
   Most of these methods propose novel building blocks for neural networks.
   These novel building blocks are specially tailored for the task of scene text recognition and can thus hardly be used in any other tasks.
   In this paper, we introduce a new model for scene text recognition that only consists of off-the-shelf building blocks for neural networks.
   Our model (KISS) consists of two ResNet based feature extractors, a spatial transformer, and a transformer.
   We train our model only on publicly available, synthetic training data and evaluate it on a range of scene text recognition benchmarks, where we reach state-of-the-art or competitive performance, although our model does not use methods like 2D-attention, or image rectification.
   Code and model are available on Github~\footnote{\url{https://github.com/Bartzi/kiss}}.
\end{abstract}

\section{Introduction}
\label{sec:introduction}

\noindent
Text is a ubiquitous entity that provides high level semantic information.
Text can not only be found in written documents such as letters or bills, but also in natural scenes on, \eg, road signs, billboards, store fronts, etc.
Being able to localize and recognize text in a natural scene image is very useful for a range of applications, such as navigation systems, content-based image retrieval, image-based machine translation, or even as support for visually impaired people.
Developing more accurate and robust recognition systems for scene text has been a widely researched challenge for several years~\cite{jaderbergSyntheticDataArtificial2014,jaderbergDeepFeaturesText2014,liaoSceneTextRecognition2019,neumannRealtimeSceneText2012,shiEndtoendTrainableNeural2016,shiASTERAttentionalScene2019}.
Accurate and robust localization and recognition of scene text is a challenging problem, because text in natural scenes appears in various forms, can be heavily distorted, blurred, or placed on challenging backgrounds.

With the emergence of deep learning methods, the performance of scene text localization and recognition systems has been boosted significantly.
Especially advances in the fields of synthetic data generation~\cite{jaderbergSyntheticDataArtificial2014}, attention modeling~\cite{DBLP:journals/corr/BahdanauCB14}, or semantic segmentation~\cite{heMaskRCNN2017} pushed the state-of-the-art significantly.
In this paper, we focus on scene text recognition, which is the task of recognizing the textual content of an image that contains a single cropped word.
In recent time, a lot of sophisticated solutions for this task have been developed.
Such sophisticated solutions include, but are not limited to: neural networks that learn to rectify an input image~\cite{shiRobustSceneText2016a,zhanESIREndToEndScene2019}, neural networks that incorporate attention mechanisms tailored specifically for the task of scene text recognition~\cite{liuCharNetCharacterAwareNeural2018,shiRobustSceneText2016a}, or methods that formulate the problem as a semantic segmentation task~\cite{liaoSceneTextRecognition2019}.
For more information on related work and how other work relates to our work, please see \autoref{sec:related_work}.

\begin{figure}
   \begin{center}
      \includegraphics[width=\linewidth]{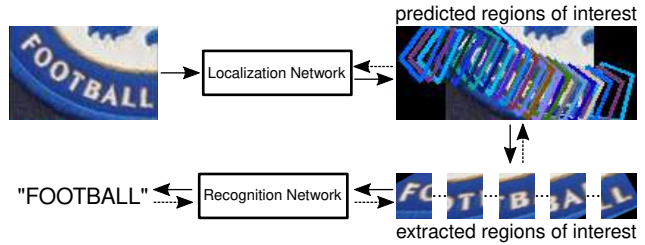}
   \end{center}
   \caption{
      Our proposed system consists of two independent neural networks that are jointly trained.
      The localization network predicts regions of interest that belong to the characters of the word in the image.
      We then use a spatial transformer to extract the regions from the image.
      The extracted regions are then fed into the recognition network that uses all extracted regions of interest to predict the textual content of the image.
      The model is end-to-end trainable (indicated by dotted lines) using only word-level annotations.
   }
   \label{fig:primer}
\end{figure}

In this paper, we propose a novel scene text recognition network that does not use any building blocks especially designed for the task of scene text recognition.
We rather use off-the-shelf components for building a neural network architecture, do not use any tricks like fine-tuning to train our network, or any non-public datasets, while still being able to reach competitive/state-of-the-art accuracy on a range of several scene text recognition benchmarks.
Our proposed model KISS (the name is based on the KISS principle~\cite{partridge2009routledge}, which says that a simple solution should be preferred over a more complex one) is straight forward and simple.
It consists of two ResNet-18 feature extractors~\cite{heDeepResidualLearning2016}, a spatial transformer~\cite{jaderbergSpatialTransformerNetworks2015} and a transformer~\cite{vaswaniAttentionAllYou2017} (see \autoref{fig:primer} for a high-level overview of our system).
Our two networks work together as a team.
The first network identifies regions of interest in the cropped word image.
Those regions of interest are predicted as affine transformation matrices that are used by the spatial transformer to crop those regions from the input image and provide them as input to the second network.
The second network now extracts features from each region of interest and uses a transformer to predict the character sequence contained in the image.
We train our model from scratch and use only publicly available synthetic datasets for training our model.
Further information on the design of our neural network can be found in \autoref{sec:method}.

We evaluate the proposed model on a variety of standard scene text recognition benchmarks and show that our model reaches competitive/state-of-the-art performance on all benchmarks.
We also conduct an ablation study, providing further insight into the importance of individual building blocks of our model.
More information on our experiments can be found in \autoref{sec:experiments}.

The contributions of this paper can be summarized as follows:
\begin{enumerate*}[label={(\arabic*)}]
   \item we propose a novel network architecture that builds on the cooperation of two independent neural networks, where one network learns by itself how to support the second network.
   \item The proposed model architecture is based on off-the-shelf neural network components, like ResNets, a spatial transformer and a transformer.
   \item Our model reaches competitive/state-of-the-art recognition accuracy on a range of benchmark datasets for scene text recognition.
   \item We show which building blocks are the most important for a state-of-the-art scene text recognition system.
\end{enumerate*}

\section{Related Work}
\label{sec:related_work}

\noindent
Over the course of time a wide variety of different methods for scene text recognition has been proposed.
Especially with the upcoming of methods based on deep learning, the performance of systems in the domain of scene text recognition was boosted significantly.
While the first methods used binarization methods and standard print ocr systems~\cite{neumannRealtimeSceneText2012}, or sliding window methods with random ferns~\cite{wangEndtoendSceneText2011} for the task of scene text recognition, more recent methods are fully based on directly applying learned feature extractors, based on deep neural networks for the task of scene text recognition.
Jaderberg~\etal propose one of the first deep learning based methods for scene text recognition~\cite{jaderbergDeepFeaturesText2014} where they use a deep neural network for character classification that is applied in a sliding window fashion, on a cropped word image.
Later, Jaderberg~\etal propose to directly apply a deep neural network on a cropped word image and predict each character with an individual softmax classifier~\cite{jaderbergSyntheticDataArtificial2014}.

Later approaches incorporate a recurrent neural network after the convolutional feature extractor, in order to utilize the capabilities of recurrent neural networks to capture sequences.
Shi~\etal propose one of the first scene text recognition models that are based on a convolutional feature extractor and a recurrent neural network~\cite{shiEndtoendTrainableNeural2016}.
He~\etal also propose a model based on a convolutional feature extractor and a recurrent neural network, but they utilize the recurrent neural network on features extracted from crops of the word image, which have been obtained in a sliding window fashion~\cite{heReadingSceneText2016}, which is different from \cite{shiEndtoendTrainableNeural2016}, where they use the complete word image as input to their convolutional feature extractor.
Compared to our method, the method by He~\etal method only uses a very simple and inflexible sliding window method that neither adapts to the number of characters in the image, nor the orientation of the characters in the image.

The idea of using a recurrent neural network to predict a character sequence has since been extended by various methods that incorporate an attention mechanism into the character sequence prediction~\cite{liShowAttendRead2019,liaoMaskTextSpotterEndtoEnd2019,liaoSceneTextRecognition2019,liuCharNetCharacterAwareNeural2018,shiRobustSceneText2016a,shiASTERAttentionalScene2019,wangFACLSTMConvLSTMFocused2019,zhanESIREndToEndScene2019}.
Liu~\etal~\cite{liuCharNetCharacterAwareNeural2018} propose a neural network that includes a spatial transformer network~\cite{jaderbergSpatialTransformerNetworks2015} which is used to put focus on the features of single characters.
They produce their character predictions using a special hierarchical character attention layer and a LSTM, following the spatial transformer and convolutional feature extractor.

The methods~\cite{shiRobustSceneText2016a,shiASTERAttentionalScene2019,zhanESIREndToEndScene2019} first utilize a spatial transformer network that predicts a thin plate spline transformation to rectify the textual content of the input image.
Then, they extract features from the rectified image with a convolutional feature extractor, followed by a recurrent neural network, with different forms of attention, for character prediction.
In~\cite{shiRobustSceneText2016a,shiASTERAttentionalScene2019} Shi~\etal utilize a sequence-to-sequence network that consists of a bidirectional LSTM as encoder and an attention guided GRU as decoder.
Zhan \& Lu~\cite{zhanESIREndToEndScene2019} rectify the input image multiple times before using a ResNet~\cite{heDeepResidualLearning2016} based sequence-to-sequence recognition network with attention.
The approach introduced by Luo~\etal~\cite{luoMORANMultiObjectRectified2019} also follows the path of image rectification before recognizing the textual content of the cropped word image, but they do not use spatial transformers to predict the necessary rectification transformation.
Instead, they predict an offset map that is applied to each pixel, achieving a differentiable sampling procedure.

Our method does not use any specialized attention methods, instead we use the spatial transformer network to generate regions of interest that (in the ideal case) contain a single and (ideally) rectified character.
We then feed the extracted region of interest into our second network that uses a transformer~\cite{vaswaniAttentionAllYou2017} to predict single characters.

In~\cite{chengAONArbitrarilyOrientedText2018} Cheng~\etal introduce an approach for the recognition of arbitrarily oriented text.
Their approach consists of a network that uses features from 4 different spatial directions, together with a special attention mechanism to predict the textual content of the image.
In contrast to our approach, this approach consists of several newly introduced and complicated building blocks that help to boost the recognition performance of the model.

Other approaches directly utilize information about the bounding box of individual characters for the generation of their predictions.
Wang~\etal~\cite{wangFACLSTMConvLSTMFocused2019} propose to use a convolutional LSTM~\cite{shiConvolutionalLSTMNetwork2015} on top of a convolutional feature extractor with a special mask based attention mechanism for the prediction of characters.
One approach by Liao~\etal formulates the task of scene text recognition as a semantic segmentation task and they create a fully convolutional network that predicts a segmentation encoding the position and class of each character~\cite{liaoSceneTextRecognition2019}.
Both methods need ground-truth bounding boxes for each character, constraining their possible extensibility, as it is expensive to obtain such a full labeling.
Our method also localizes individual characters, but without the need for labeled character bounding boxes, making it simpler to apply our method to new datasets.
Another approach by Liao~\etal~\cite{liaoMaskTextSpotterEndtoEnd2019} utilizes a sequence-to-sequence network with attention guided GRU.
They utilize a two dimensional feature map as input to their attention guided decoder, which is similar to the work by Shi~\etal~\cite{shiASTERAttentionalScene2019}, but they do not use a rectification network. 

Wang~\etal~\cite{wangSimpleRobustConvolutionalAttention2019} also propose a simple network that only consists of a feature extractor and a transformer for scene text recognition.
In our work, we also use a transformer, but we train it in a different way.

A method that highly influenced our work, is the work by Bartz~\etal~\cite{bartzSEESemiSupervisedEndtoEnd2018}.
They propose a neural network architecture for the task of end-to-end scene text recognition and show that a model can be trained for text localization and recognition, solely under the supervision of a text recognition based objective.
We build on their approach and make fundamental changes to the overall network structure.

\section{Method}
\label{sec:method}

\noindent
Our network consists of two parts.
We call the first part \emph{localization network} and the second part \emph{recognition network}.
Our localization network consists of a ResNet based feature extractor and a spatial transformer that produces the input to the second network and allows us to train localization network and recognition network at the same time.
The recognition network consists of a ResNet based feature extractor and a transformer.
The recognition network takes regions of interest as input and produces a character sequence as output.
In \autoref{fig:network} we provide a structural overview of the proposed network.
In this section we present each building block in detail and also introduce a novel method that is necessary to successfully train the whole network.

\subsection{Localization Network}
\label{subsec:localization_network}

\begin{figure}[t]
   \begin{center}
      \includegraphics[width=\linewidth]{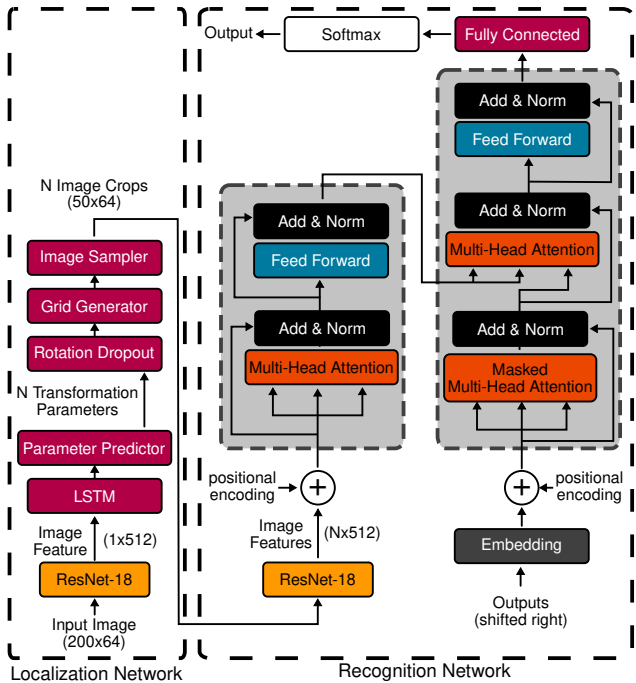}
   \end{center}
   \caption{
      The network architecture proposed in this work consists of two sub-networks.
      Each network first extracts features using a ResNet-18 feature extractor.
      The first network then produces $N$ regions of interest, using a recurrent spatial transformer.
      Each of these $N$ regions is then used as input to the recognition network.
      The recognition network then uses a transformer to predict the textual content of the input image, after extracting features with the ResNet-18 feature extractor.
   }
   \label{fig:network}
\end{figure}

\noindent
The task of the localization network is to predict regions of interest that (ideally) contain a character.
The localization network is trained by the recognition network, which in turn is trained using word-level annotations.
Our localization network consists of a convolutional feature extractor and a recurrent spatial transformer, as can be seen on the left hand side of \autoref{fig:network}.

\paragraph{Feature Extractor}
We use a convolutional neural network based on the ResNet~\cite{heDeepResidualLearning2016} architecture, as our convolutional feature extractor.
We choose to use a ResNet based feature extractor because ResNet does not suffer from the vanishing gradient problem, as much, as other network architectures, such as, VGG~\cite{simonyanVeryDeepConvolutional2015}, or Inception~\cite{DBLP:conf/cvpr/SzegedyLJSRAEVR15} do.

\paragraph{Spatial Transformer}
Following the convolutional feature extractor, we use a spatial transformer~\cite{jaderbergSpatialTransformerNetworks2015}.
The task of the spatial transformer is to extract regions of interest, that are likely to contain a character.
In order to extract these regions of interest the spatial transformer consists of three different parts:
\begin{enumerate*}[label={\arabic*.}]
 	\item a transformation predictor,
 	\item a grid generator, and
 	\item a differentiable image sampler.
\end{enumerate*}

The transformation parameter predictor uses the features that have been extracted by the feature extractor and predicts a series of $N$ affine transformation matrices $A^n_{\theta}$.
The parameter $N$ is the maximum number of characters our system shall predict.
In all of our experiments we follow the setting of Jaderberg~\etal~\cite{jaderbergSyntheticDataArtificial2014} and set $N$ to \num{23}.
Those affine transformation matrices can be used to perform affine transformations like scaling, rotating, translating or skewing the pixels of the input image.
Each transformation matrix $n$ with $n \in [0, N)$ consists of six parameters $\theta$ that are predicted using the extracted features:
\begin{equation}
	A^n_{\theta} = \begin{bmatrix}
		\theta^n_1 & \theta^n_2 & \theta^n_3 \\
		\theta^n_4 & \theta^n_5 & \theta^n_6
	\end{bmatrix}.
\end{equation}
Each of these matrices $A^n_{\theta}$ is predicted using a recurrent neural network, in our case a LSTM~\cite{hochreiterLongShortTermMemory1997}.
At each timestep $n$ of our LSTM we use a fully connected layer with \num{6} neurons to predict the transformation matrix $A^n_{\theta}$.
We use rotation dropout~\cite{bartzSEESemiSupervisedEndtoEnd2018}, which works like dropout~\cite{srivastavaDropoutSimpleWay2014} but only drops the parameters $\theta^n_2$ and $\theta^n_4$, of the transformation matrix $A^n_{\theta}$.
These parameters are responsible for performing rotations on the input image.
During all of our experiments we set the dropout rate to \num{0.05}.

The next part of the spatial transformer is the grid generator.
The grid generator uses the predicted transformation matrices to create a set of $N$ sampling grids with width $w_o$ and height $h_o$.
Those sampling grids provide the coordinates $u^n_i, v^n_j$ (with $i \in [0, w_o)$ and $j \in [0, h_o)$) that are used to sample from the input image, using the differentiable image sampler.
The size of each grid defines the size of the output of the spatial transformer.
This allows us to use images of different input sizes in the localization network, while maintaining a fixed input size for the recognition network.
The coordinates $u^n_i, v^n_j$ are produced by multiplying the coordinates $x_i,y_j$ ($x,y = [-1, \ldots, 1]$) of an evenly spaced grid $G$ with one of our predicted transformation matrices $A^n_{\theta}$:
\begin{equation}
	\begin{pmatrix}
		u^n_i \\ v^n_j
	\end{pmatrix} = 
	\begin{bmatrix}
		\theta^n_1 & \theta^n_2 & \theta^n_3 \\
		\theta^n_4 & \theta^n_5 & \theta^n_6
	\end{bmatrix}
	\begin{pmatrix}
		x_i \\ y_j \\ 1
	\end{pmatrix}.
\end{equation}

\noindent
Previous work~\cite{bartzSEESemiSupervisedEndtoEnd2018,shiRobustSceneText2016a,shiASTERAttentionalScene2019,zhanESIREndToEndScene2019} proposes to initialize the parameter predictor for the affine transformation matrix $A^n_{\theta}$ in such a way that a box spanning a large portion of the image is predicted.
They claim this is necessary, because otherwise the network would not converge.
We, on the contrary, found that randomly initializing the parameter predictor works very well and even helps to speed up the convergence of the model.
However, we found that extra regularization is necessary, for a network with a randomly initialized parameter predictor to converge.
Without extra regularization, the network tends to predict transformation parameters that produce regions of interest that are far outside of the bounds of the input image.
Since regions out of the bounds of the image only contain zeros, the network can not learn from these examples anymore, keeping the network from converging.
In order to encourage the network to not predict such transformation parameters, we propose an \emph{out-of-image regularizer} that is directly applied on the sampling grid, generated by the grid generator.
Since the coordinates $x$ and $y$ of the grid $G$ are in the interval $[-1, \ldots, 1]$, thus spanning the entire image before the application of our predicted transformation matrix, we know that each value for $u$ and $v$ should exactly be within this interval, in order to be inside of the image.
We can now create a regularizer that penalizes each predicted value $u$ and $v$ if they are not in the interval:
\begin{equation}
	reg(c) = \lvert min(c + 1, 0) \rvert + max(c - 1, 0).
\end{equation}
With $c$ being any value from $u$ and $v$.
The value of $reg(u)$ and $reg(v)$ can be added to the overall loss of the network, as shown in \autoref{eq:loss}.

The last part of the spatial transformer is the image sampler~\cite{jaderbergSpatialTransformerNetworks2015}.
The image sampler uses the predicted sampling grids to crop the corresponding pixels from the input image in a differentiable way.
Since the sampling points of the predicted sampling grid do not perfectly align with the discrete grid of pixels in the input image, bilinear sampling is used.
The image sampler produces $N$ different output images $O^n$, where each output image has the same size, as the sampling grid, used for sampling the pixels.
Each pixel of the output image $O^n$ is then defined to be:
\begin{equation}
	O^n_{ij} = \sum^{H}_h \sum^{W}_w I_{hw} max(0, 1 - \lvert u^n_i - h \rvert)max(0, 1 - \lvert v^n_j - w \rvert).
\end{equation}
Where $I$ is the input image with width $W$ and height $H$.

\subsection{Recognition Network}

\noindent
The recognition network is the most important part of our proposed system.
The recognition network takes the regions of interest that have been extracted by the localization network, or any other method, and predicts the character sequence contained in these  region images.
The right hand side of \autoref{fig:network} provides a schematic overview of the recognition network.
This network is trained using word-level annotations and can propagate its gradients to an upstream network, in our case the localization network.
The recognition network consists of a convolutional feature extractor and a transformer that predicts the sequence of characters in the image.
The convolutional feature extractor is, as the feature extractor of the localization network, based on the ResNet-18 architecture.

Each region of interest that is used as input to the recognition network is processed independently by the feature extractor.
The extracted features are used as input to the transformer and a sequence with $N$ elements is formed, based on the output of the localization network.
For our transformer, we follow the design introduced in \cite{vaswaniAttentionAllYou2017}.
A transformer consists of an encoder and a decoder.
The encoder takes the extracted features of our $N$ regions of interest, adds a positional encoding to the features of each region of interest $n \in [0, N)$, performs self attention on the feature map and produces an encoded feature representation.
The decoder takes the last predicted character, or a begin-of-sequence token as input, embeds the input into a high dimensional space, applies positional encoding, and masked self-attention.
The masked self-attention is followed by multi-head attention that decides on which parts of the encoded feature vector to attend to, which is in turn followed by a feed forward layer and a classifier.
In the following we will briefly describe the most important parts of the transformer, namely the positional encoding, multi-head attention and the combination of encoder and decoder.

\paragraph{Positional Encoding}

The positional encoding is used to add some information about the order of the sequence of features to the transformer.
This is necessary, since the transformer does not use any recurrence relations that can implicitly encode an order, hence an explicit encoding is necessary.
We follow \cite{vaswaniAttentionAllYou2017} and use sine and cosine functions for the positional encoding:
\begin{align}
	PE_{(n, 2d)} &= sin(n/ e^{(\frac{2d}{d_{model}} \cdot log(10000))}), \\
	PE_{(n, 2d+1)} &= cos(n/ e^{\frac{2d}{d_{model}} \cdot log(10000))}).
\end{align}
Where $n \in [0, N)$, $d_{model}$ is the dimensionality of the model, and $d \in [0, d_{model})$ is the current dimension.

\paragraph{Multi-Head Attention}

The most important part of the transformer is the usage of attention.
The advantages of using only attention instead of a recurrence relation with attention is the lower computational complexity and that the input sequence can be processed in parallel.
A transformer is also easier to train, compared to a RNN, since there is no recurrence relation and the gradient can flow better to each input.
We follow \cite{vaswaniAttentionAllYou2017} and use the same formulation of multi-head attention as they do.
Multi-head attention consists of multiple scaled dot-product attention blocks that each operate on a different subset of available channels.
Since each of these attention heads operates on a subset of the available information, each head becomes an expert for a specific kind of feature.
For further information about the inner structure of the used multi-head attention, please refer to \cite{vaswaniAttentionAllYou2017}.
Multi-head attention is used in several forms and at different positions of the transformer.
The self-attention block in the encoder uses multi-head attention.
Here the encoder decides which features to attend to for each region of interest.
The decoder uses masked self-attention.
The self-attention in the decoder is masked because we do not want the decoder to have a look into the future, but base its attention focus on the past outputs.
On top of the masked self-attention, we use multi-head attention to combine encoder and decoder.
The multi-head attention where encoder and decoder are combined works like a typical encoder-decoder attention mechanism~\cite{DBLP:journals/corr/BahdanauCB14} and is guided by the output of the masked self-attention in the decoder.
Besides the attention layers, the transformer also contains a feed forward layer to warp the attended feature maps, independently for each region of interest.
Each of these transformer building blocks (as can be seen in \autoref{fig:network}) form one transformer layer, which can be stacked multiple times.
Note that we use a single transformer layer instead of a stack of multiple transformer layers.
In our ablation study (see \autoref{subsec:ablation_study}) we show that stacking multiple transformer layer does not increase performance for a model like ours.

\paragraph{Training Objective}

Following the transformer we use a fully connected layer, which is accompanied by a softmax classifier.
The softmax classifier predicts a probability distribution over the possible character classes we trained our model to distinguish.
We use 95 different classes, including digits, case-sensitive letters from a to z, 32 symbols, and a blank symbol.
We train our model using softmax cross-entropy loss.
Besides the cross-entropy loss, we also add the localizer specific regularization terms, as discussed in \autoref{subsec:localization_network}, to the overall training objective:

\begin{align}
	\mathcal{L}_{network} &= CrossEntropy(F_{rec}(F_{loc}(I)), G_t) \\ 
	\mathcal{L} &= \mathcal{L}_{network} + \sum_i reg(u_i) + \sum_j reg(v_j).
	\label{eq:loss}
\end{align}
Where $I$ is the input image, $G_t$ is the ground-truth annotation for the input image $I$, while $F_{loc}$ and $F_{rec}$ denote the localization and recognition network, respectively.


	
\section{Experiments}
\label{sec:experiments}

\noindent
We evaluate our model on a range of scene text recognition benchmarks.
In this section, we first introduce the benchmarks that we performed our experiments on.
Second, we describe further implementation details and hyperparameters used for training and testing our model.
Then we show the results of our best performing model on the benchmark datasets.
Last but not least, we conduct an ablation study to show the impact of changes to hyper parameters and also how changes to the network structure and training methodology impact the performance of our model.

\begin{figure}[t]
    \begin{center}
        \includegraphics[width=\linewidth]{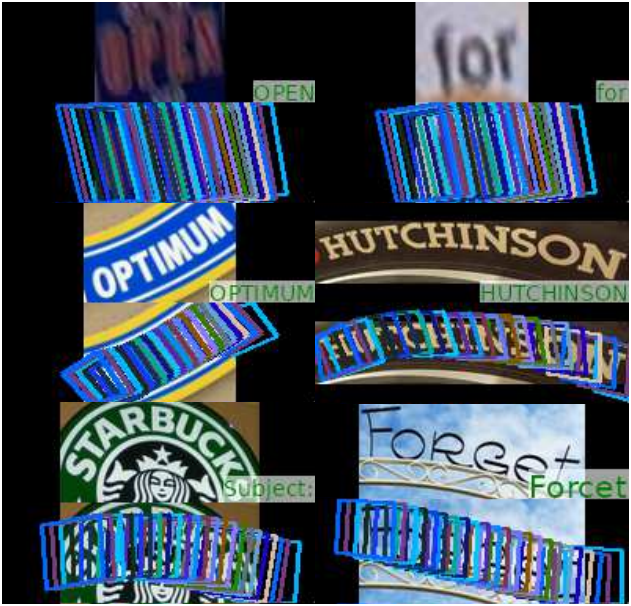}
    \end{center}
    \caption{
        Qualitative evaluation of our approach.
        We always show the input image with the predicted word and also the predicted regions of interest, below the input image.
        The last row shows two failure cases of our model.
    }
    \label{fig:qualitative_eval}
\end{figure}

\subsection{Datasets}
\label{subsec:datasets}

We train our model only on publicly available synthetic datasets.
Our model is evaluated on each benchmark dataset without further fine-tuning.
For training we use the following datasets: MJSynth~\cite{jaderbergSyntheticDataArtificial2014}, SynthText in the Wild~\cite{guptaSyntheticDataText2016a}, and SynthAdd~\cite{liShowAttendRead2019}.
We evaluate our model on the following datasets: ICDAR 2013 (IC13)~\cite{karatzasICDAR2013Robust2013}, ICDAR2015 (IC15)~\cite{karatzasICDAR2015Competition2015}, IIIT5K-Words (IIIT5K)~\cite{mishraSceneTextRecognition2012}, Street View Text (SVT)~\cite{wangEndtoendSceneText2011}, Street View Text Perspective (SVTP)~\cite{quyphanRecognizingTextPerspective2013}, and CUTE80 (CUTE)~\cite{risnumawanRobustArbitraryText2014}.
In the following, we shortly introduce each dataset:

\begin{description}[leftmargin=0cm]
    \item[Training Datasets] We train our model on publicly available datasets, in order to allow a fair comparison of our model to other models.
    We use MJSsynth~\cite{jaderbergSyntheticDataArtificial2014}, which consists of 9 million synthetic word images, SynthText in the Wild~\cite{guptaSyntheticDataText2016a}, which consists of 8 million word images, and SynthAdd~\cite{liShowAttendRead2019}, which consists of 1.6 million word images.
    \item[ICDAR 2013 (IC13)~\textnormal{\cite{karatzasICDAR2013Robust2013}}] consists of \num{1095} word images for evaluation. Those images contain focused scene text. For a fair comparison with other approaches, we remove all images with non-alphanumeric characters, leaving us with \num{1015} images for evaluation.
    \item[ICDAR 2015 (IC15)~\textnormal{\cite{karatzasICDAR2015Competition2015}}] consists of \num{2077} word images for evaluation. The images have been captured using Google Glass without the intent to capture scene text, hence they are severely distorted, or blurred. For fair comparison, we also evaluate on the ICDAR2015-1811 (IC15-1811) subset, which only contains alpha numeric characters.
    \item[IIIT5K-Words (IIIT5K)~\textnormal{\cite{mishraSceneTextRecognition2012}}] consists of \num{3000} word images for evaluation. Most of the word images are horizontal text, but several images also contain curved text instances.
    \item[Street View Text (SVT)~\textnormal{\cite{wangEndtoendSceneText2011}}] consists of \num{647} word images for evaluation, most of which are horizontal text lines, but many images are severely blurred or of bad quality.
    \item[Street View Text Perspective (SVTP)~\textnormal{\cite{quyphanRecognizingTextPerspective2013}}] consists of \num{645} word images for evaluation. These images have been collected from Google StreetView and contain images with a high rate of distortions.
    \item[CUTE80 (CUTE)~\textnormal{\cite{risnumawanRobustArbitraryText2014}}] consists of \num{288} word images for evaluation. The images are mostly of high quality but contain a lot of curved text instances.
\end{description}

\subsection{Implementation Details}
\label{subsec:implementation_details}

\noindent
We implemented our model\footnote{implementation and model are available on Github: \url{https://github.com/Bartzi/kiss}} using Chainer~\cite{tokuiChainerNextGenerationOpen2015}, and based our implementation of the transformer for Chainer on~\cite{opennmt}.
We perform all of our experiments on a NVIDIA GTX 1080Ti GPU.
We use RAdam~\cite{liuVarianceAdaptiveLearning2019} as optimizer, and set the initial learning rate to $10^{-4}$.
We multiply the learning rate by \num{0.1} at each new epoch.
We train our model for about 3 epochs, use a batch size of $32$, and resize the input images to $200\times64$, while maintaining the aspect ratio.
The output of the localization network is set to be image crops of size $50\times64$, which is also used as input size for the recognition network.
Our model is trained from scratch with randomly initialized weights and we use GroupNormalization~\cite{DBLP:conf/eccv/WuH18} instead of BatchNormalization~\cite{ioffeBatchNormalizationAccelerating2015b} throughout the network.
We follow \cite{vaswaniAttentionAllYou2017} when setting the hyperparameters of the transformer, but we use only one layer of encoder and decoder instead of six.
During training we use gradient clipping in the localizer to avoid divergence of the network.
We use some data augmentation during training of our model and augment \SI{40}{\percent} of the input images, by randomly resizing them, adding some blur, and adding some extra distortion.
During test time, following~\cite{wangSimpleRobustConvolutionalAttention2019}, we also perform some data augmentation.
We rotate each input image by $\pm 5$ degrees, if the width is \num{1.3} times larger than the height, otherwise we rotate the image by $\pm 90$ degrees.
We then put all rotated images and the original image into the network.
We use the prediction with the highest average score for each character as output of our network.

\subsection{Comparison with State-of-the-art}
\label{subsec:comparison_with_state_of_the_art}

\begin{table*}[t]
    \begin{center}
    \begin{tabular}{r | c | c | c | c | c | c | c}
    \hline
    Method & IC13 & IC15 & IC15-1811 & IIIT5K & SVT & SVTP & CUTE \\
    \hline\hline
    Jaderberg~\etal 2014~\cite{jaderbergSyntheticDataArtificial2014} & \num{90.8} & - & - & - & \num{80.7} & - & - \\
    Shi~\etal 2016~\cite{shiEndtoendTrainableNeural2016} & \num{89.6} & - & - & \num{81.2} & \num{82.7} & - & - \\
    Shi~\etal 2016~\cite{shiRobustSceneText2016a} & \num{88.6} & - & - & \num{81.9} & \num{81.9} & - & - \\
    Liu~\etal 2018~\cite{liuCharNetCharacterAwareNeural2018} & \num{91.1} & \textit{74.2} & - & \num{92.0} & \num{85.5} & \num{78.9} & - \\
    Cheng~\etal 2018~\cite{chengAONArbitrarilyOrientedText2018} & - & \num{68.2} & - & \num{87.0} & \num{82.8} & \num{73.0} & \num{76.8} \\
    Bai~\etal 2018~\cite{baiEditProbabilityScene2018} & \textit{94.4} & - & \num{73.9} & \num{88.3} & \num{85.9} & - & - \\
    Shi~\etal 2019~\cite{shiASTERAttentionalScene2019} & \num{91.8} & - & \num{76.1} & \num{93.4} & \num{89.5} & \num{78.5} & \num{79.5} \\
    Wang~\etal 2019~\cite{wangFACLSTMConvLSTMFocused2019} & - & - & - & \num{90.5} & \num{82.2} & - & \num{83.3} \\
    Li~\etal 2019~\cite{liShowAttendRead2019} & \num{91.0} & \num{69.2} & - & \num{91.5} & \num{84.5} & \num{76.4} & \num{83.3} \\
    Luo~\etal 2019~\cite{luoMORANMultiObjectRectified2019} & \num{92.4} & \num{68.8} & - & \num{91.2} & \num{88.3} & \num{76.1} & \num{77.4} \\
    Liao~\etal 2019~\cite{liaoSceneTextRecognition2019} & \num{91.5} & - & - & \num{91.9} & \num{86.4} & - & \num{79.9} \\
    Wang~\etal 2019~\cite{wangSimpleRobustConvolutionalAttention2019} & \num{92.0} & \textbf{74.8} & \textit{79.1} & \textit{94.2} & \num{89.0} & \num{81.7} & \num{83.7} \\
    Zhan \& Lu 2019~\cite{zhanESIREndToEndScene2019} & \num{91.3} & - & \num{76.9} & \num{93.3} & \textit{90.2} & \num{79.6} & \num{83.3} \\
    Liao~\etal 2019~\cite{liaoMaskTextSpotterEndtoEnd2019} & \textbf{95.3} & - & \num{77.3} & \num{93.9} & \textbf{90.6} & \textit{82.2} & \textit{87.8} \\
    \hline\hline
    KISS (Ours) & \num{93.1} & \textit{74.2} & \textbf{80.3} & \textbf{94.6} & \num{89.2} & \textbf{83.1} & \textbf{89.6}
    \end{tabular}
    \end{center}
    \caption{
        Results of our model compared to other proposed models. Numbers in \textbf{bold font} and \textit{italic font} indicate the best and second-best performance, respectively.
        We only report results that are obtained without any lexicon.
        It is possible to see that our model outperforms every other model on irregular text datasets, showing the general robustness to distortions of our model.
    }
    \label{tab:state_of_the_art_comparison}
\end{table*}

\noindent
We compare our model with several other methods on the datasets introduced in \autoref{subsec:datasets}.
For a fair comparison we always show the results that were obtained using only synthetic data and without an ensemble of multiple models (where known).
We show the results of our best performing model (exactly the model described in \autoref{sec:method}) in \autoref{tab:state_of_the_art_comparison}.
Our method, although without any components tailored to the task of scene text recognition, reaches state-of-the-art results on four of the seven benchmarks and reaches competitive results on the other three benchmarks.
We note that we can not evaluate our model on the entire ICDAR2015 dataset, since this dataset contains a range of characters we did not train our model for.
This accounts for about \SI{1.8}{\percent} of the images in the ICDAR2015 dataset.
Even though we did not use building blocks specifically tailored to the task of scene text recognition, we are still able to achieve state-of-the-art results.
We are still wondering, why this is the case.
In our ablation study (\autoref{subsec:ablation_study}) we try to show which building blocks of our network boost the performance of our network the most, but it still does not answer the question, why our method outperforms other methods.
In our future work, we will investigate this further.
In \autoref{fig:qualitative_eval}, we show some qualitative results of our approach.
It is possible to see that our model works well on distorted and heavily blurred text.
The failure cases show that our model is not able to handle heavily curved text well, this might be because the training dataset lacks this kind of image for training.

\subsection{Ablation Study}
\label{subsec:ablation_study}

\begin{table*}[t]
    \begin{center}
    \begin{tabular}{r | c | c | c | c | c | c | c}
    \hline
    Variation & IC13 & IC15 & IC15-1811 & IIIT5K & SVT & SVTP & CUTE \\
    \hline\hline
    SynthText only & 90.4 & 65.8 & 71.1 & 89.2 & 82.4 & 73.3 & 74.7 \\
    Balanced dataset & 91.8 & 72.5 & 77.6 & 93.1 & 86.7 & 77.3 & 89.2 \\
    Transformer Localizer & 91.0 & 70.6 & 76.3 & 92.5 & 87.9 & 78.0 & 85.8 \\
    Softmax Recognizer & 86.0 & 54.7 & 59.4 & 82.2 & 74.6 & 60.1 & 68.4 \\
    $\text{Transformer}^2$ w/o Augmentation & 93.0 & 72.8 & 75.3 & 90.8 & 88.3 & 79.8 & 85.4 \\
    Recognition Network only & 90.9 & 67.5 & 73.6 & 91.5 & 85.0 & 74.6 & 78.1 \\
    Without Augmentation & 92.7 & 72.9 & 75.4 & 90.8 & 88.3 & 79.2 & 87.2\\
    Recognizer Optimizer & 92.5 & 73.4 & 77.0 & 94.5 & \textbf{90.0} & 79.8 & 87.2\\
    \hline\hline
    KISS & \textbf{93.1} & \textbf{74.2} & \textbf{80.3} & \textbf{94.6} & 89.2 & \textbf{83.1} & \textbf{89.6} \\
    \end{tabular}
    \end{center}
    \caption{
        Results of our ablation experiments. \textbf{Bold font} indicates the best performance for the corresponding dataset.
    }
    \label{tab:ablation_study}
\end{table*}

\noindent
Besides the comparison with state-of-the-art methods, we also perform a range of experiments where we make changes to several building blocks of our network or changes to the way we train the network.
With these experiments we want to determine the performance impact of those changes and also which building blocks are important for a state-of-the-art scene text recognition model.
The results of our ablation experiments can be seen in \autoref{tab:ablation_study}.
We performed the following ablation experiments:
\begin{description}[leftmargin=0cm]
    \item[Training Data]
        In the first experiment, we do not use all datasets as described in \autoref{subsec:datasets}, but only the SynthText dataset~\cite{guptaSyntheticDataText2016a}, which includes about \SI{41}{\percent} of the overall training images.
        With this experiment we want to determine the impact a reduced amount of training data has on our model.
        The result (SynthText only) shows that reducing the amount of training data has a drastic impact on performance, indicating that using only one of the synthetic datasets is not enough to train a well balanced model.
        We also perform an experiment where we balance the training dataset.
        Here, we take all three synthetic datasets and only keep a maximum of \num{200000} samples per word-length.
        This leaves us with about \SI{60}{\percent} of the overall training images.
        The result for this experiment (Balanced dataset) shows that this does not decrease the performance of the model as drastically, as using only the SynthText dataset does, which shows that all three datasets together contain a well balanced range of examples and allow the model to better generalize to real-world examples.
    \item[Changes to the network architecture]
        In this series of experiments, we exchange several building blocks of our network architecture and observe the impact of these changes.
        First, we exchange the LSTM in the localization network with a transformer.
        Here we want to answer two questions:
        \begin{enumerate*}[label={(\arabic*)}]
            \item is it possible to train a transformer without immediately providing the labels in the forward pass?
            \item Does a transformer based localizer increase the performance of the model, especially on heavily distorted images?
        \end{enumerate*}
        The results (Transformer Localizer) show that it is indeed possible to train a transformer without the need to provide labels in the forward pass, but we note that it takes significantly longer than training a RNN, because the full sequence has to be created step by step.
        We also see from the results that a transformer based localizer does not increase the performance of the model, hence a transformer does not provide much value at this stage of the network.
        We note that we are currently not enabling the transformer to attend to a two dimensional feature map.
        Doing so could improve the results we could achieve using a localization network with a transformer, we leave that open for future work.
        The second experiment we conduct, is to exchange the transformer in the recognition network with a plain softmax classifier, as it is used in~\cite{jaderbergSyntheticDataArtificial2014}.
        The result (Softmax Recognizer) shows that using a sequence to sequence model with attention (in our case a transformer) is indeed one of the most important building blocks for a successful text recognition model.
        Next, we increase the number of transformer layers of the transformer in the recognition network.
        The result ($\text{Transformer}^2$ w/o Augmentation) shows that this change leads to similar performance than without this change, hence it is sufficient to stay with the simpler approach.
        The last experiment is to train the model without the localization network on regularly cropped regions of interest.
        The result (Recognition Network only) shows that our model performs very good on datasets that contain focused scene text.
        On datasets that do not contain focused scene text or a lot of curved text instances, this model is outperformed by our proposed model.
        This shows that a careful selection of regions of interest is important for the network to succeed, which confirms the observations by~\cite{shiRobustSceneText2016a,shiASTERAttentionalScene2019,zhanESIREndToEndScene2019}, where they argue that rectification is beneficial for the task of scene text recognition.
    \item[Changes to the Training Method]
        In our last series of experiments, we experiment with different ways of training the model.
        First, we train the model without data augmentation.
        From the result (Without Augmentation), we can see that data augmentation is crucial for achieving a good result on several datasets, while on other datasets, augmentation does not seem to have a large impact.
        This shows that our current augmentation strategy is suited very well for the data in some datasets and also supports the findings of our first ablation experiments, showing that  training data is a crucial factor for a very good performance.
        Another change we made to the training methodology is to add an extra optimizer that only trains the recognition network on regularly cropped slices of a word image.
        The rationale behind this idea is, that this extra training might help the recognition network to perform better, as it sees more diverse input data and also better cropped regions of interest, since the start of the training.
        However, the results (Recognizer Optimizer) show that this is not the case.
        Using this training methodology also does not decrease the performance of the model by a large margin.
\end{description}

\section{Conclusion}
\label{sec:conclusion}

In this paper, we have presented KISS, a neural network for scene text recognition that only consists of off-the-shelf neural network building blocks.
Our model architecture is simple and can be trained end-to-end from scratch.
We also presented methods to successfully train a model with a spatial transformer, but without the need for careful parameter initialization.
In our experiments, we showed that our model reaches competitive/state-of-the-art results compared to other approaches.
Our results show that it is possible to reach state-of-the-art performance, using only off-the-shelf neural network components.
In our ablation study we showed that using a sequence-to-sequence model with attention and the availability of diverse training data are the most important factors for creating a state-of-the-art model for scene text recognition.
But, these results still do not answer the question why our method performs better than others.
In the future, we would like to investigate this further and show which modules are the most important ones for the task of scene text recognition.


{\small
\bibliographystyle{ieee_fullname}
\bibliography{egbib}
}

\end{document}